\documentclass{article}

\usepackage{arxiv}

\usepackage[utf8]{inputenc} % allow utf-8 input
\usepackage[T1]{fontenc}    % use 8-bit T1 fonts
\usepackage{hyperref}       % hyperlinks
\usepackage{url}            % simple URL typesetting
\usepackage{booktabs}       % professional-quality tables
\usepackage{amsfonts}       % blackboard math symbols
\usepackage{nicefrac}       % compact symbols for 1/2, etc.
\usepackage{microtype}      % microtypography
\usepackage{lipsum}
\usepackage{natbib}
\usepackage{amsmath}
\usepackage{listings}
\usepackage{wrapfig}
\usepackage{graphicx}
\graphicspath{ {images/} }
\usepackage{caption}
\usepackage{subcaption}
\usepackage{algpseudocode}
\usepackage{algorithm}
\title{PyDEns: a \textsc{Python}-framework for solving differential equations with neural networks}

\author{
  Alexander Koryagin \\
  Data Analysis Center\\
  Gazpromneft\\
  Saint Petersburg \\
   \And
Roman Khudorozkov \\
  Data Analysis Center\\
  Gazpromneft\\
  Saint Petersburg \\
  \And
 Sergey Tsimfer\\
  Data Analysis Center\\
  Gazpromneft\\
  Saint Petersburg \\
}

\begin{document}
\maketitle

\begin{abstract}
Recently, a lot of papers proposed to use neural networks to approximately solve partial differential equations (PDEs). Yet, there has been a lack of flexible framework for convenient experimentation. In an attempt to fill the gap, we introduce a \textsc{PyDEns}-module open-sourced on \textsc{GitHub}. Coupled with capabilities of \textsc{BatchFlow}, open-source framework for convenient and reproducible deep learning, \textsc{PyDEns}-module allows to 1) solve partial differential equations from a large family, including \textit{heat equation} and \textit{wave equation} 2) easily search for the best neural-network architecture among the zoo, that includes \textsc{ResNet} and \textsc{DenseNet} 3) fully control the process of model-training by testing different point-sampling schemes. With that in mind, our main contribution goes as follows: implementation of a ready-to-use and open-source numerical solver of PDEs of a novel format, based on neural networks.
\end{abstract}

\section{Introduction}
It is widely known that neural networks are universal approximators of functions (\cite{Cybenko1989}). \cite{Lagaris1998} proposed to use this property of neural networks to solve partial differential equations. In doing so, they formed a trial function using neural network and ran an optimization procedure so that the differential equation be as valid as possible for the trial. In order to form the loss-function, the authors manually computed high-order (corresponds to the degree of equation) derivatives of the network output w.r.t to its inputs and weights. Later, \cite{Lagaris2000} extended their approach on domains with more complex boundaries. The main idea was to form a trial function using two neural networks. The first one solves the equation inside the domain while the second one binds the boundary conditions.

After mentioned papers, the domain of research took a long pause. In recent years, however, following the development of convenient frameworks for automatic differentiation (e.g. TensorFlow and PyTorch), more and more papers (\cite{Berg2018}, \cite{Nabian2018}, \cite{Aradi2018}) attempt to solve differential equations using neural networks. To our understanding, all of the modern approaches can be traced back to a \textsc{DeepGalerkin}-method introduced by \cite{Sirignano2018}. Essentially, \cite{Sirignano2018} rediscovered the approach of \cite{Lagaris1998} and used the power of modern automatic differentiation to test the approach at a larger scale, using deep neural networks for solving tasks at high dimensions.

In this paper, we wrap up the progress in the domain of solving PDEs with neural networks in a \textsc{PyDEns} \textsc{python}-module. The module is available for deep learning community on \textsc{GitHub} for usage and further improvement. Our goal is to simplify the experimentation in the emerging area of research. In doing so, we rely upon several pillars. First of all, a user should have an opportunity to set up complex problems in several lines of clear code. In that way, \textsc{PyDEns} allows to set up a PDE-problem from a wide class, including but not limited to \textit{heat equation} and \textit{wave equation}. Secondly, it is important not to impose constraints on the choice of neural network-architecture. With \textsc{PyDEns}, a user can either (i) choose a network from a zoo of implemented architectures, including \textsc{ResNet}, \textsc{DenseNet}, \textsc{MobileNet} and others or (ii) build a complex neural network from scratch in a line of code, using convolutions and fully-connected layers. Lastly, a user has complete control over point-sampling scheme. One can form batches of training data using a large family of probabilistic distributions on the domain of PDE, e.g. truncated gaussian, uniform and exponential distributions or even mixtures of those three.

The rest of the paper is organized as follows: in the next section we briefly review \textsc{DeepGalerkin}-model as it was introduced in the paper of \cite{Sirignano2018} and present our modifications to the algorithm. In section (\ref{probsetup}) we explain in detail how to set up a \textsc{PyDEns} \textsc{Python}-model for solving a particular PDE-problem. In section (\ref{bestpractice}) we give a few recommendation regarding (i) the choice of architecture, (ii) sampling schemes and (iii) batch size. We conclude by talking about further development of our approach and explaining how our work relates to oil-gas industry.

\section{Original Deep Galerkin and our modifications}
\subsection{\textit{Original Deep Galerkin}}
Given evolution PDE of the form
\begin{gather}
    \frac{\partial u}{\partial t} + \mathcal{L}u(t, x) = 0,\quad (t, x) \in [0, T] \times \Omega,\quad \Omega  \subset \mathcal{R}^d,
    \label{sys:1} \\
    u(t=0, x) = u_0(x),\label{sys:2} \\
    u(t, x) = g(t, x),\quad x \in \sigma \Omega, \label{sys:3} \\
    \notag \mathcal{L}\ \textrm{is spatial differential operator}.
\end{gather}
\cite{Sirignano2018} seek to approximate the solution $u(t, x)$ using neural network $net(t, x; \theta)$. In doing so, they minimize the following objective:
\begin{gather}
    J(\theta) = \left\| \frac{\partial \rm{net}}{\partial t}(t, x; \theta) + \mathcal{L} \rm{net}(t, x; \theta) \right\|^2_{(t, x) \in [0, T] \times \Omega,\nu_1}  + \left\|\rm{net}(t, x;\theta) - g(t, x) \right\|^2_{(t, x) \in [0, T] \times \sigma\Omega,\nu_2} + \label{obj} \\
    \notag + \left\|\rm{net}(0, x;\theta) - u_0(x)\right\|^2_{x \in \Omega,\nu_3}.
\end{gather}
In other words, the neural network $net(x; \theta)$ is fitted to validate all components (\ref{sys:1})-(\ref{sys:3}) of PDE, both the equation itself and boundary/initial conditions. Importantly, each term of objective (\ref{obj}) is given by an integral along its respective domain w.r.t. some measure $\nu_i$. Naturally, these integrals are rarely tractable. Hence, \cite{Sirignano2018} propose to replace the integrals on its sample-counterparts and use stochastic gradient descent. All in all, the optimization process looks as follows:
\begin{algorithm}
    \caption{Deep Galerkin-algorithm}
    \label{algo:original}
    Fit $net(t; x; \theta)$ to approximately solve the PDE.
    \begin{algorithmic}[1]
    \Procedure{DG}{$\nu_1, \nu_2, \nu_3$}
    \Repeat
        \State
        Form batches of points $b_1$, $b_2$, $b_3$ from $[0, T] \times \Omega$, $[0, T] \times \sigma \Omega$, $\{0\} \times \Omega$. Draw from distributions $\nu_1$, $\nu_2$, $\nu_3$.
        \State
        Estimate the objective (\ref{obj}) using sampled batches:
        \begin{gather*}
            \widehat{J}(\theta) = \sum\limits_{(t_i, x_i) \in  b_1}
            \left( \frac{\partial \rm{net}}{\partial t}(t_i, x_i; \theta) + \mathcal{L} \rm{net}(t_i, x_i; \theta)      \right)^2 + \sum\limits_{(t_i, x_i) \in b_2} \left(\rm{net}(t_i, x_i;\theta) - g(t_i, x_i) \right)^2 + \\ +
            \sum\limits_{(0, x_i) \in b_3} \left(\rm{net}(0, x_i;\theta) - u_0(x_i)\right)^2.
        \end{gather*}
        \State
        Take a step of gradient descent:
        \begin{gather*}
            \theta_{n+1} = \theta_n - \alpha_n \nabla_{\theta} \widehat{J}(\theta_n)
        \end{gather*}
    \Until convergence.
    \EndProcedure
\end{algorithmic}
\end{algorithm}

Note that this algorithm allows only for \textit{approximate} binding of initial and boundary conditions by means of incorporating additional terms into objective (\ref{obj}).

\subsection{\textit{Problem description}}
With \textsc{PyDEns} one can solve almost any imaginable PDE. Still, the model focuses on \textit{evolution equations with time up to the second order}:
\begin{equation}
F\left(u; t, x_1,\dots, x_{n-1};
\frac{\partial u}{\partial t},\frac{\partial^2 u}{\partial t^2};
\dots, \frac{\partial u}{\partial x_i}, \dots
\frac{\partial^2 u}{\partial x_i x_j}, \dots
\frac{\partial^3 u}{\partial x_i x_j x_k},\dots
\right) = 0.
 \label{eq:PDE}
\end{equation}
where $F$ is an arbitrary function, composed from well-known operations including $\sin$, $\cos$, <<$+$>>, <<$-$>> as well as differential operators $\frac{\partial{}}{\partial x_i}, \frac{\partial{}}{\partial t}$. The domain of the problem is given by $n$-dimensional rectangle $\Omega \in \mathcal{R}^n$:
$$
\Omega = [\Omega_0^1, \Omega_1^1] \times \dots \times [\Omega_0^n, \Omega_1^n].
$$
To form a well-posed equation of evolution one also needs boundary conditions:
\begin{equation}
    \left. u(x) \right\rvert_{x \in \partial\Omega} = g(x), \label{eq:BC}
\end{equation}
and initial state along with evolution rate of the system\footnote{needed \textbf{only} if $\frac{\partial^2 u}{\partial{t}^2}$ is present in the equation}:
\begin{equation}
    u(x_1, \dots, x_{n-1}, t_0) = u_0(x_1, \dots, x_{n-1}), \label{eq:IC0}
\end{equation}
\begin{equation}
    \left. \frac{\partial u(x_1, \dots, x_{n-1}, t)}{\partial t} \right\rvert_{t=t_0} = u_0'(x_1, \dots, x_{n-1}). \label{eq:IC1}
\end{equation}

\subsection{\textit{Introducing ansatz: binding boundary and initial conditions}}
When working on PDE-problems posed on rectangular domains, the out-of-the box approach of \textsc{PyDEns}-module uses \textit{ansatz} for exact binding of initial/boundary conditions\footnote{the original algorithm from \cite{Sirignano2018} can also be easily implemented}:
\begin{equation}\label{ansatz}
    A[\text{net}](x, t; \theta) = \text{mutliplier}(x, t; \theta) * \text{net}(x, t; \theta) + \text{addendum}(x, t; \theta).
\end{equation}
In other words, the solution to PDE is approximated by a transformation of a neural network-output rather than $net(x, t;\theta)$ itself. Form (\ref{ansatz}) ensures exact binding of initial/boundaries conditions (\ref{eq:BC})-(\ref{eq:IC1}) whenever the following holds:
\begin{gather*}
    \text{mutliplier}(x, t=0; \theta) = 0,\quad \text{addendum}(x, t=0; \theta) = u_0(x).\\
    \text{mutliplier}'_t(x, t=0; \theta) = 0,\quad \text{addendum}'_t(x, t=0; \theta) = u_0'(x)\\
    \left.\text{mutliplier}(x, t; \theta)\right\rvert_{x \in \partial\Omega} = 0,\quad \left. \text{addendum} \right\rvert_{x \in \partial\Omega} = g(x).
\end{gather*}
In all, the modified algorithm of \textsc{PyDEns}-module looks as follows (\ref{algo:modified}) - compare this to algorithm (\ref{algo:original}):

\begin{algorithm}
    \caption{Modified Deep Galerkin-algorithm}
    \label{algo:modified}
    Fit
    \[A[\text{net}](x, t; \theta) = \text{mutliplier}(x, t; \theta) * \text{net}(x, t; \theta) + \text{addendum}(x, t; \theta).\]
    to approximately solve the PDE.
    \begin{algorithmic}[1]
    \Procedure{DG}{$\nu_1$}
    \Repeat
        \State
        Form batch of points $b_1$ from $[0, T] \times \Omega$. Draw from distribution $\nu_1$.
        \State
        Estimate the objective (\ref{obj}) using sampled batches:
        \begin{gather*}
            \widehat{J}(\theta) = \sum\limits_{(t_i, x_i) \in  b_1}
            \left( \frac{\partial \rm{A}}{\partial t}(t_i, x_i; \theta) + \mathcal{L} \rm{A}(t_i, x_i; \theta) \right)^2.
        \end{gather*}
        \State
        Take a step of gradient descent:
        \begin{gather*}
            \theta_{n+1} = \theta_n - \alpha_n \nabla_{\theta} \widehat{J}(\theta_n)
        \end{gather*}
    \Until convergence.
    \EndProcedure
\end{algorithmic}
\end{algorithm}

\section{Problem setup and configuration}\label{probsetup}
\subsection{\textsc{BatchFlow} and config-dictionaries}
In order to overcome many obstacles related to the training of neural networks, \textsc{PyDEns} relies on \textsc{BatchFlow}-framework. The main purpose of \textsc{BatchFlow} is to allow for creation of reproducible pipelines for deep learning. Most importantly, \textsc{BatchFlow} allows to set up a neural network model for training in a simple way, using \textit{configuration dictionary}:
\begin{lstlisting}
    config = {"loss": "mse",
              "optimizer": "Adam",
              "body": {...}}
\end{lstlisting}
The go-to approach of \textsc{BatchFlow} for creating a neural network-architecture from scratch is to use complex block \textsc{\texttt{conv\_block}}. The block uses \textit{string layout} for defining the network as a sequence of layers:
\begin{lstlisting}
    body = {"layout": "faR fa fa+ f",
            "units": [10, 25, 10, 1],
            "activation": [tf.nn.tanh, tf.nn.tanh, tf.nn.tanh]}
\end{lstlisting}
In the code section above layout "faR fa fa+ f" stands for \textbf{f}ully connected network with 3$(=4 - 1)$ hidden layers, hyperbolic tangent-\textbf{a}ctivations and one \textsc{ResNet}-like skip connection: symbols \textbf{R} and \textbf{+} stand for the start and the end of the connection respectively.
In the same manner, \textsc{PyDEns} can be configured to solve a particular PDE-problem via \textit{pde-key} of the configuration-dict. The next section explains in detail how to correctly set up \textit{pde-key}.
\subsection{Configuring \textsc{PDE}-key}\label{generalcase}
We start filling up the pde-key by specifying the dimensionality of the equation:
\begin{lstlisting}
    pde = {"n_dims": 2}
\end{lstlisting}
The next step is to define the equation itself using the \textit{key form}. This key sets up differential operator $F$ of equation (\ref{eq:PDE}) as a \textsc{Python}-callable. To define the callable we use the language of mathematical tokens. The list of tokens includes names like $\sin, \cos, exp$ and differentiation-token $D(\cdot, x)$. The first step is to add a set of tokens to the current namespace via `add tokens` function:
\begin{lstlisting}
    from pydens import add_tokens
    add_tokens()
\end{lstlisting}
We can go on with defining the equation. For purposes of demonstration let us setup the equation
$$
\frac{\partial^2 u(x, y)}{\partial x^2} + \frac{\partial^2 u(x, y)}{\partial y^2}  = Q(x, y).
$$
known as \textit{Poisson equation}:
\begin{lstlisting}
    Q = lambda x, y: 5 * sin(np.pi * (x + y))
    form = lambda u, x, y: D(D(u, x), x) + D(D(u, y), y) - Q(x, y)
    pde.update(form=form)
\end{lstlisting}
As you can see, the usage of tokens is rather straightforward. Note that token $D(\cdot, x)$ can be chained allowing for creation of higher-order derivatives.
Default value for domain $\Omega$ of the equation is unit $n$-dimensional rectangle $[0, 1]^n$. To change it, one can pass 'domain' key in PDE setup dictionary:
\begin{lstlisting}
    pde.update(domain=[[0, 1], [0, 1])
\end{lstlisting}
To finish configuration of the PDE-problem, we must supply the boundary condition:
\begin{lstlisting}
    pde.update({"boundary_condition": 1})
\end{lstlisting}
It is not difficult to define a more complex boundary condition:
\begin{lstlisting}
    pde.update({"boundary_condition": lambda x, y: exp(y) * sin(x) + cos(y)})
\end{lstlisting}
\subsection{Configuring the rest of the model}\label{rest_model}
We go on with configuring our model and define a simple feed-forward architecture with three hidden layers to solve the PDE at hand:
\begin{lstlisting}
    body = {"layout": "fa fa fa f",
            "units": [15, 25, 15, 1],
            "activation": [tf.nn.tanh]*3}
\end{lstlisting}

We can now assemble model-configuration dictionary using previously defined \textsc{pde} and \textsc{body}-subconfigs and adding mse-loss function:
\begin{lstlisting}
    model_config = {"pde": pde,
                    "body": body,
                    "loss": "mse"}
\end{lstlisting}
Finally, we need to specify strategy of generating points from the domain. In this example we simply use uniform distribution over unit square in $\mathcal{R}^2$:

\begin{lstlisting}
    from pydens import NumpySampler
    sampler = NumpySampler("uniform", dim=2)
\end{lstlisting}

To learn more on how to create complex distributions, check out our \url{https://github.com/analysiscenter/batchflow/blob/master/examples/tutorials/07_sampler.ipynb}.

\begin{figure}
    \centering
    \begin{minipage}{.5\textwidth}
      \centering
      \includegraphics[width=1.0\linewidth]{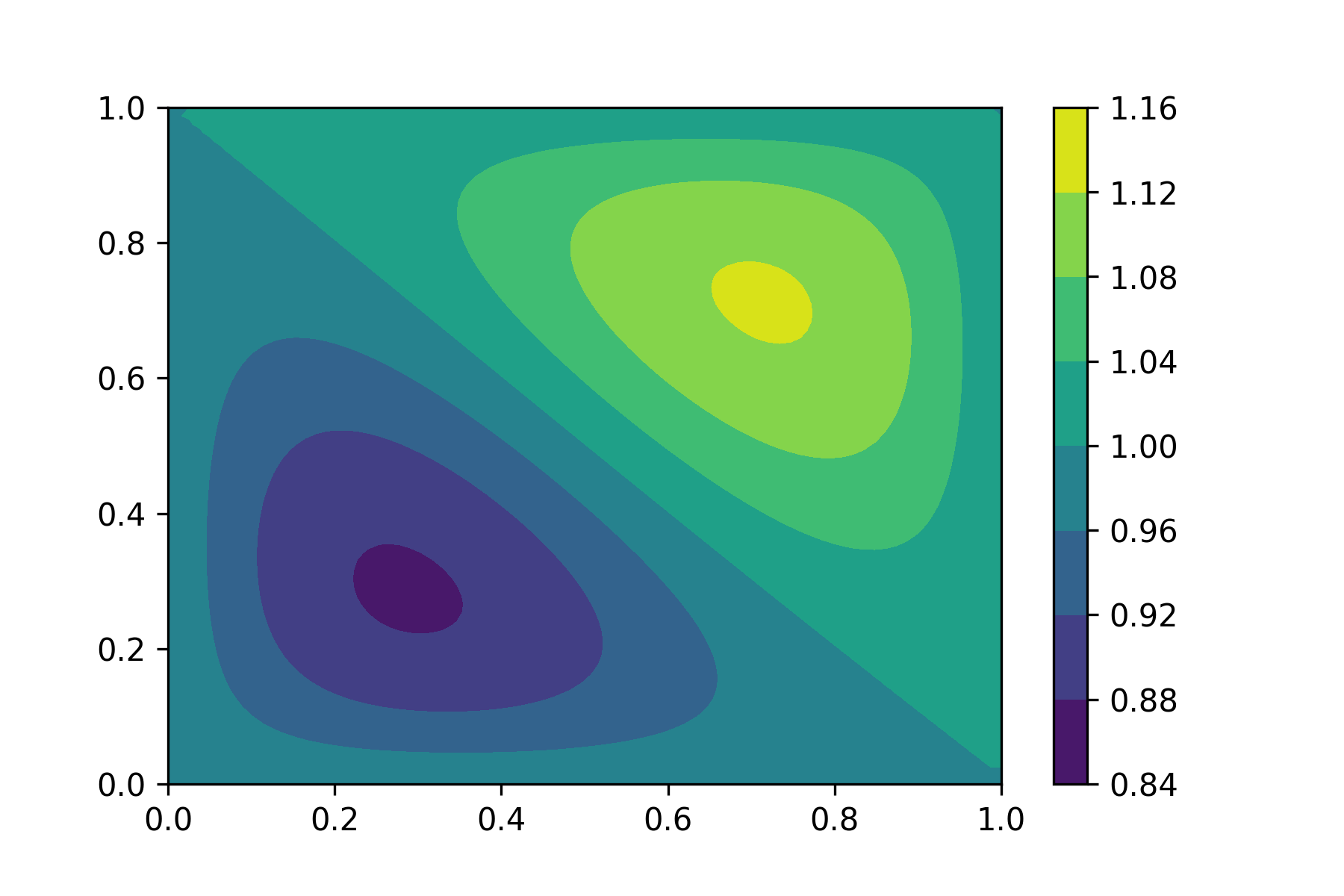}
      \captionof{figure}{Neural network approximation: $2d$-view}
      \label{fig:loss_p}
    \end{minipage}%
    \begin{minipage}{.5\textwidth}
      \centering
      \includegraphics[width=1.0\linewidth]{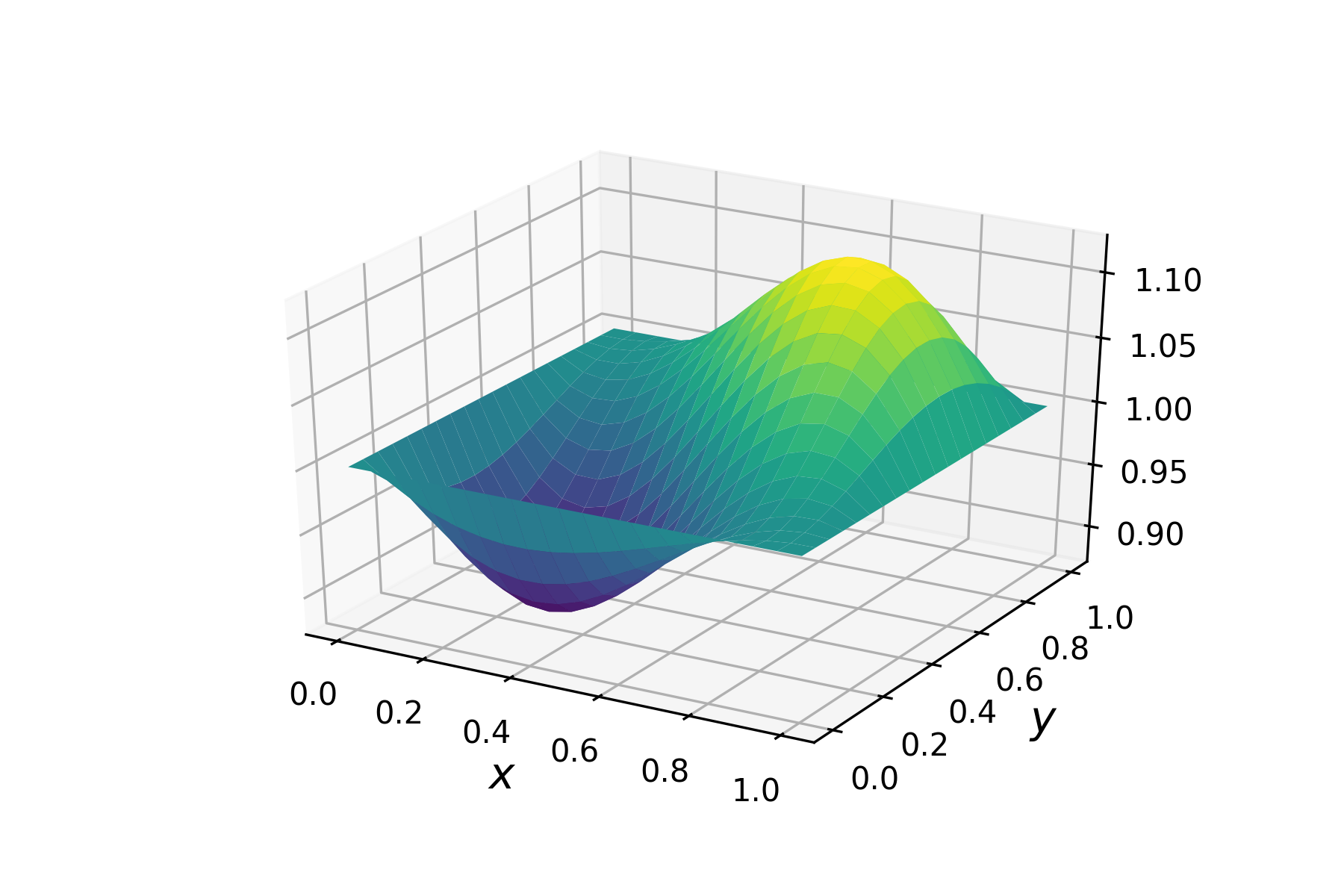}
      \captionof{figure}{Neural network approximation: $3d$-view}
      \label{fig:sol_p}
    \end{minipage}
    \centering
    \includegraphics[width=8.5cm, height=4.5cm]{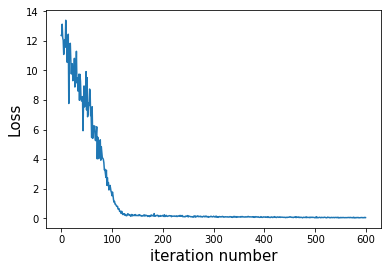}
    \captionof{figure}{Loss against training time}
    \label{fig:loss_p_}
\end{figure}

All that is left now is to train configured model in order to minimize error-function. For that we only need to wrap up the model-config in a model-instance and run the fitting procedure:
\begin{lstlisting}
    from pydens import DGSolver
    dg = DGSolver(config)
    dg.fit(sampler=sampler, batch_size=200, n_iters=1000)
\end{lstlisting}

Change of loss-function \eqref{obj} is illustrated on Figure \ref{fig:loss_p}. The fact that it achieves zero demonstrates that neural network approximates solution of the given PDE on desired domain. From graph of approximation, shown on Figure \ref{fig:sol_p}, we can easily verify that boundary conditions are satisfied.

In the next section we demonstrate how one can configure \textsc{PyDEns} to solve a more complex problem.

\subsection{Heat equation}\label{heat}
Lets turn our attention to heat equation of the form
\begin{gather*}
    \frac{\partial u(x, y, t)}{\partial t} - \frac{\partial^2 u(x, y,  t)}{\partial x^2} - \frac{\partial^2 u(x, y, t)}{\partial y^2} = 5 xy(1-x)(1-y) \cos{(\pi(x+y))}, \\
    u(x, y, 0) = xy(1-x)(1-y).
\end{gather*}

To communicate this problem to the model, we use following syntax:

\begin{lstlisting}
    form = lambda u, x, y, t: (D(u, t) - D(D(u, x), x) - D(D(u, y), y)
                                   - 5 * x * y * (1 - x) * (1 - y) * cos(np.pi * (x + y)))
    ic = lambda x, y: x * y * (1 - x) * (1 - y)

    pde = {"form": form,
           "rhs": rhs,
           "initial_condition": ic}
\end{lstlisting}

Since the problem at hand is not as simple as the previous one, we need more sophisticated neural network as approximator:

\begin{lstlisting}
    body = {"layout": "faR fa fa+ f",
            "units": [10, 25, 10, 1],
            "activation": [tf.nn.tanh] * 3}
\end{lstlisting}

\begin{wrapfigure}{r}{0.5\textwidth}
  \vspace{-20pt}
  \centering
  \includegraphics[width=.5\textwidth]{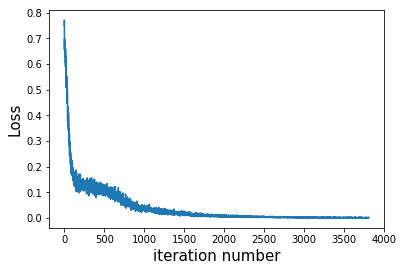}
  \captionof{figure}{Heat equation. Loss against training time}
  \vspace{-60pt}
  \label{fig:loss_w}
\end{wrapfigure}

\begin{figure}[b]
  \centering
  \includegraphics[width=.99\textwidth]{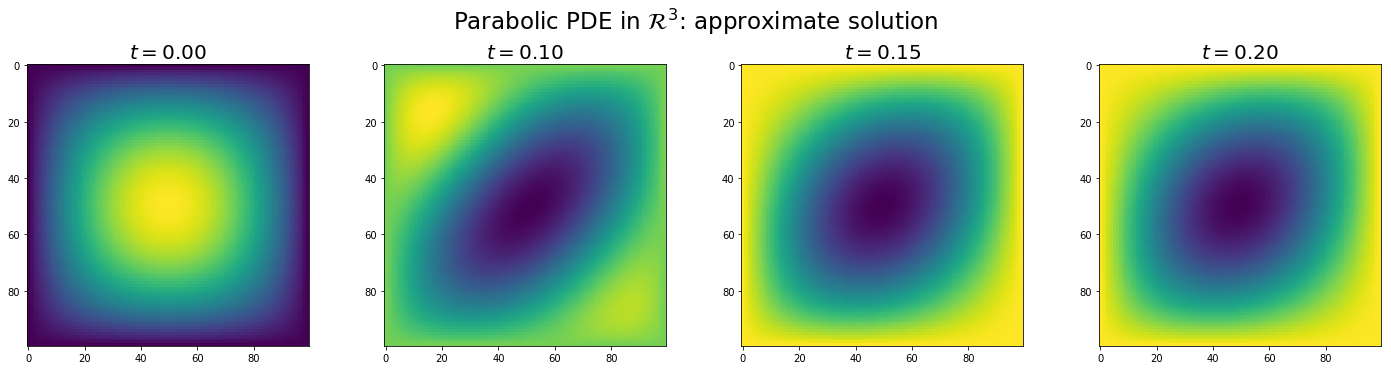}
  \captionof{figure}{Heat equation. Neural network approximation}
  \label{fig:sol_w}
\end{figure}

Letter 'R' in the layout convention stands for beginning of residual connection, while plus sign for its ending with summation. Rest of the solving pipeline is pretty much the same: uniform sampling over unit square and batch size of 200 points are used. Results are demonstrated on Figures \ref{fig:loss_w}, \ref{fig:sol_w}.

\section{Choice of hyperparameters}\label{bestpractice}
In this section we discuss choice of important hyperparameters, namely, model architecture, number of points in every training batch, use of different sampling schemes and so on.

\subsection{Model architecture}
In previous examples we almost exclusively used fully-connected layers and residual connections. For harder equation in hand, especially with fast changing solutions, it is recommended to use gated connections, such as ones introduced in \cite{hochreiter1997long}, \cite{cho2014learning}. Number of layers and their respective size depends mostly on the dimensionality of PDE.

\subsection{Batch size}
Amount of points in each training batch is of utmost importance. If it is not big enough, then the estimation of true gradient is too inaccurate and leads to poor performance. On the other hand, if there are enough points to cover big part of the domain, then each training step does not account for any region-specific information, which also leads to low quality of the resulting model. The optimal middle-ground can be found by looking at graph of loss function during training: big oscillations usually speak for too low of a batch size, while unsatisfactory value of loss after plateauing can be viewed as a sign of too many points in the batch, provided that neural network is expressive enough.

\subsection{Sampling schemes}
Another way of communicating region-specific information to the model is by carefully choosing sampling strategy. Analogous to classical methods of solving PDE's, we can sample points near provided boundary or initial condition during the first few iterations of model fitting. In the same manner, we can use different samplers during all of the training time: each of them would concentrate on some sufficiently small region of the initial domain, so that the model is trained to work better in this exact part of $\Omega$. It translates to code in a straight-forward manner:

\begin{lstlisting}
    sampler_1 = NumpySampler(...)
    sampler_2 = NumpySampler(...)

    dg.fit(sampler=sampler_1, ...)
    dg.fit(sampler=sampler_2, ...)
\end{lstlisting}

In applications, we usually want to know solution only for narrow location. This is often the case for hydrodynamic modelling in oil-gas. Subsequent usage of
\begin{itemize}
    \item initial sampler near boundary or initial conditions
    \item domain-wide sampler or combination of region-specific samplers to achieve small value of loss on the domain as a whole
    \item sampler, concentrated around region of interest
\end{itemize}
allows to fine-tune model to perform best at the desired location.

\section{Conclusion}
We have presented \textsc{PyDEns} open-source \textsc{python}-module based on work of \cite{Sirignano2018}. The framework allows to train neural networks to approximately solve second-order PDE's of evolution. In order to set up a PDE-problem one only has to (i) communicate the problem itself in a \textsc{python}-dictionary (ii) set up a neural network-architecture using easy-to-comprehend layouts and (iii) prepare a point-sampling scheme using algebra of samplers, combining base \textsc{numpy}-distributions in mixtures and product-distributions. In all, the framework allows for more convenient experimentation in emerging domain of solving PDEs with neural networks.

In further work we plan to focus on (i) incorporating uncertainty into equation-inputs (for instance, its coefficients) in spirit of \cite{Aradi2018} and (ii) making coefficients of the equation trainable. This is of special importance in relation to oil-gas industry, as it sets a path for a drastically new approach for (i) predicting the evolution of oil-gas fields under uncertain geological properties and (ii) solving filtration problem (or other problems of inverse modeling).
\bibliographystyle{plainnat}
\bibliography{references}

\begin{thebibliography}{9}
\providecommand{\natexlab}[1]{#1}
\providecommand{\url}[1]{\texttt{#1}}
\expandafter\ifx\csname urlstyle\endcsname\relax
  \providecommand{\doi}[1]{doi: #1}\else
  \providecommand{\doi}{doi: \begingroup \urlstyle{rm}\Url}\fi

\bibitem[Al-Aradi et~al.(2018)Al-Aradi, Correia, Naiff, Jardim, and
  Saporito]{Aradi2018}
Ali Al-Aradi, Adolfo Correia, Danilo Naiff, Gabriel Jardim, and Yuri Saporito.
\newblock Solving nonlinear and high-dimensional partial differential equations
  via deep learning, 2018.

\bibitem[Berg and Nystr\"{o}m(2018)]{Berg2018}
Jens Berg and Kaj Nystr\"{o}m.
\newblock A unified deep artificial neural network approach to partial
  differential equations in complex geometries.
\newblock \emph{Neurocomputing}, 317:\penalty0 28--41, nov 2018.
\newblock \doi{10.1016/j.neucom.2018.06.056}.
\newblock URL \url{https://doi.org/10.1016/j.neucom.2018.06.056}.

\bibitem[Cho et~al.(2014)Cho, van Merrienboer, Gulcehre, Bougares, Schwenk, and
  Bengio]{cho2014learning}
Kyunghyun Cho, Bart van Merrienboer, Caglar Gulcehre, Fethi Bougares, Holger
  Schwenk, and Yoshua Bengio.
\newblock Learning phrase representations using rnn encoder-decoder for
  statistical machine translation.
\newblock \emph{arXiv preprint arXiv:1406.1078}, 2014.

\bibitem[Cybenko(1989)]{Cybenko1989}
G.~Cybenko.
\newblock Approximation by superpositions of a sigmoidal function.
\newblock \emph{Mathematics of Control, Signals, and Systems}, 2\penalty0
  (4):\penalty0 303--314, dec 1989.
\newblock \doi{10.1007/bf02551274}.
\newblock URL \url{https://doi.org/10.1007/bf02551274}.

\bibitem[Hochreiter and Schmidhuber(1997)]{hochreiter1997long}
Sepp Hochreiter and J{\"u}rgen Schmidhuber.
\newblock Long short-term memory.
\newblock \emph{Neural computation}, 9\penalty0 (8):\penalty0 1735--1780, 1997.

\bibitem[Lagaris et~al.(1998)Lagaris, Likas, and Fotiadis]{Lagaris1998}
I.E. Lagaris, A.~Likas, and D.I. Fotiadis.
\newblock Artificial neural networks for solving ordinary and partial
  differential equations.
\newblock \emph{{IEEE} Transactions on Neural Networks}, 9\penalty0
  (5):\penalty0 987--1000, 1998.
\newblock \doi{10.1109/72.712178}.
\newblock URL \url{https://doi.org/10.1109/72.712178}.

\bibitem[Lagaris et~al.(2000)Lagaris, Likas, and Papageorgiou]{Lagaris2000}
I.E. Lagaris, A.C. Likas, and D.G. Papageorgiou.
\newblock Neural-network methods for boundary value problems with irregular
  boundaries.
\newblock \emph{{IEEE} Transactions on Neural Networks}, 11\penalty0
  (5):\penalty0 1041--1049, 2000.
\newblock \doi{10.1109/72.870037}.
\newblock URL \url{https://doi.org/10.1109/72.870037}.

\bibitem[Nabian and Meidani(2018)]{Nabian2018}
Mohammad~Amin Nabian and Hadi Meidani.
\newblock A deep neural network surrogate for high-dimensional random partial
  differential equations.
\newblock \emph{CoRR}, abs/1806.02957, 2018.
\newblock URL \url{http://arxiv.org/abs/1806.02957}.

\bibitem[Sirignano and Spiliopoulos(2018)]{Sirignano2018}
Justin Sirignano and Konstantinos Spiliopoulos.
\newblock {DGM}: A deep learning algorithm for solving partial differential
  equations.
\newblock \emph{Journal of Computational Physics}, 375:\penalty0 1339--1364,
  dec 2018.
\newblock \doi{10.1016/j.jcp.2018.08.029}.
\newblock URL \url{https://doi.org/10.1016/j.jcp.2018.08.029}.

\end{thebibliography}

\end{document}